\documentclass[conference]{IEEEtran}
\IEEEoverridecommandlockouts
\usepackage{amsmath,amssymb,amsfonts}
\usepackage{algorithmic}
\usepackage{algorithm}
\usepackage{xcolor}
\usepackage{array}
\usepackage[caption=false,font=normalsize,labelfont=sf,textfont=sf]{subfig}
\usepackage{textcomp}
\usepackage{stfloats}
\usepackage{url}
\usepackage{verbatim}
\usepackage{graphicx}
\usepackage{cite}
\usepackage{stmaryrd}
\usepackage [mathscr] {eucal}
\usepackage{flushend} 

\begin{document}

\title{Concurrent Haptic, Audio, and Visual Data Set During Bare Finger Interaction with Textured Surfaces}


\author{Alexis Devillard$^{*1}$, Aruna Ramasamy$^{*2,3}$,\\
Damien Faux$^{3}$, Vincent Hayward$^{3}$~\IEEEmembership{Life Fellow,~IEEE}, and Etienne Burdet$^{1}$
\thanks{$*$ Equal contribution. This work was funded in part by the EC grant INTUITIVE (ITN 861166).}
\thanks{$^1$ Department of Bioengineering, Imperial College of Science, Technology and Medicine, London, UK. \{awd20, e.burdet\}@imperial.ac.uk}
\thanks{$^2$ École Normale Supérieure, CNRS, Laboratoire des Systèmes Perceptifs, 24 rue Lhomond, 75005, Paris, France}
\thanks{$^3$ Actronika, 157 Boulevard Macdonald, 75019 Paris, France. \{aruna.ramasamy, damien.faux, vincent.hayward\}@actronika.com}
\thanks{Corresponding authors: Alexis Devillard, Aruna Ramasamy}}



\maketitle

\begin{abstract}
Perceptual processes are frequently multi-modal. This is the case of haptic perception. Data sets of visual and haptic sensory signals have been compiled in the past, especially when it comes to the exploration of textured surfaces. These data sets were intended to be used in natural and artificial perception studies and to provide training data sets for machine learning research. These data sets were typically acquired with rigid probes or artificial robotic fingers. Here, we collected visual, auditory, and haptic signals acquired when a human finger explored textured surfaces. We assessed the data set via machine learning classification techniques. Interestingly, multi-modal classification performance could reach 97\% when haptic classification was around 80\%.

\end{abstract}


\section{Introduction}
\IEEEPARstart{I}{t} is well known that human perception is often multi-sensory where different sources of information accessed through different sensory modalities are merged and integrated by the brain. This integration process is thought to increase the robustness of the perception of the properties of objects in the face of uncertainty, to resolve ambiguities, and to contribute to the perceptual stability of sensory scenes~\cite{ernst_merging_2004,bertelson_psychology_2004,munoz_multisensory_2012,durrant-whyte_sensor_1988}.

In haptics, the senses which most frequently contribute to the perception of the mechanical properties of surfaces are vision and audition, in addition to touch. It stands to reason that the collection of data sets for use in perceptual studies, artificial perception, and machine learning research should include rich data from these three sensory modalities and their sub-modalities.

In vision, because an image depends on illumination and thus impacts shading even at small scales, richer information would be available from stereoscopic rather than cyclopean viewing~\cite{anderson_stereoscopic_1999}. In audition, the mechanical interaction in sliding contacts frequently produces audible sounds~\cite{akay_acoustics_2002}, which can be an important source of information~\cite{kassuba_multisensory_2013}. In tool-mediated touch, the nature of the probe, the kinematics of scanning, and the effort applied all contribute to perceptual information.

Previous works focused on recording images of textures and associated acceleration signals produced by a rigid probe while exploring a rough surface~\cite{culbertson_generating_2013,culbertson_one_2014,strese_haptic_2014,zheng_deep_2016,strese_content-based_2017}. A few studies have also recorded the sound produced during sliding
interactions\cite{burka_proton_2016}. These studies, however, relied on rigid probes to record haptic signals such as load and  acceleration. Yet other studies have recorded the load and the vibrations generated in the substrate when a bare finger slide on a textured surface~\cite{wiertlewski_1f_2011,platkiewicz_recording_2014} or skin vibrations during sliding~
\cite{tanaka_finger-mounted_2012,kirsch_harnessing_2021}. To our knowledge, however, no rich data set is available with synchronised visual, audio, and haptic signals arising from a bare finger exploring surfaces. Such data set would be useful to conduct the aforementioned perceptual, artificial perception, and machine learning studies.

This observation motivated us to create a multi-modal data set comprising the signals created when a bare finger explored varied textured surfaces. The measured signals were stereoscopic images of the surface, the position and speed of the fingertip in images coordinates, the load applied by the finger, the sound emitted, and the vibrations propagating in the finger. While the option of employing an artificial finger to carry out the recordings, e.g \cite{xu_tactile_2013,camillieri_artificial_2017,friesen_bioinspired_2015}, was available, we preferred producing the recordings with real fingers.

To illustrate the opportunities afforded by this new data set, we compared the accuracy of classifiers trained with different combinations of data from different modalities to identify different surfaces. In the foregoing, Section \ref{Methods} describes the data acquisition apparatus, the acquisition protocol, and the surfaces employed, Section~\ref{Results} describes the performance of various machine learning classifiers, and Section~\ref{Discussion} discusses the results and suggests further work.

\section{Methods} \label{Methods}

\begin{figure}[ht!]
\centering
\includegraphics[width=0.45\textwidth]{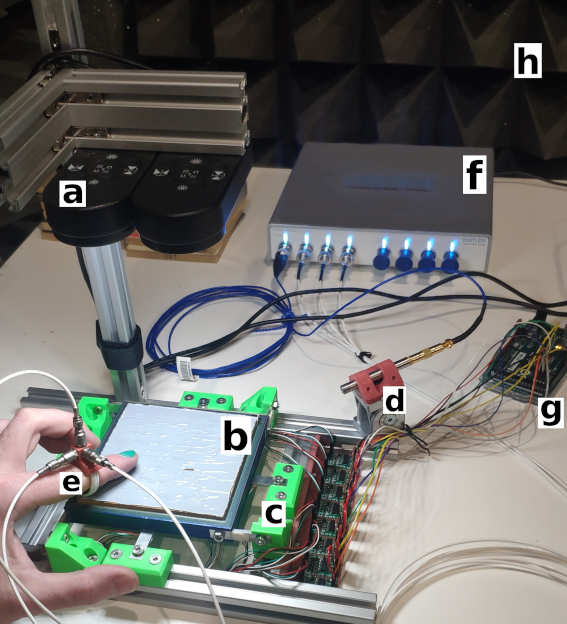}
\caption{Setup: (a)~4K cameras above a textured surface. (b)~load sensing system (c). (d)~Directional microphone with adjustable support. (e)~Accelerometers affixed to the finger touching the surface. (f) Microphone and accelerometers connected to signal conditioning unit. (g)~load sensing system connected to an Arduino. Setup installed in a sound-proofed room.}
\label{fig:setup_pic}
\end{figure}

We recorded the interactions of one participant's finger sliding against ten different surfaces. For each surface, the recorded signals comprised high resolution visual data and five trials of one minute of haptic and audio data. The signals were: \par
\begin{itemize}
    \item \textbf{Visual}: Stereoscopic images.
    \item \textbf{Audio}: Sound produced by the finger-surface interaction.
    \item \textbf{Vibration}: Vibrations propagating in the finger along three-axes of acceleration.
    \item \textbf{Kinematics}: Planar position and speed of the fingertip.
    \item \textbf{Force}: Normal and tangential force components.
\end{itemize}

\subsection{Physical Setup}
The setup, see Fig.~\ref{fig:setup_pic}, comprised two 4K cameras, a directional microphone, a set of load cells placed under the touched texture and three 1-axis accelerometers mounted on a ring. These sensors were all connected to a single computer. The recordings were conducted in a soundproof room to focus on the sound generated by the finger interacting with the material and to limit background noise.

\subsubsection{Cameras}
The cameras (Model QXC-700, Thustar, Dongguan, China) were fitted with integrated close-range lenses. They focused on the samples at a distance of 15~cm with an horizontal offset of 5~cm to provide stereoscopic views. The cameras were mounted on an aluminium profile structure to ensure positional stability. 
\paragraph{Images}
Each surface was photographed with 3840$\times$2160 resolution. To increase the variability in each image class, the position and intensity of the light sources were changed for each image of the same surface.
\paragraph{Position and velocity}
The cameras were also used to capture the position of the fingertip interacting with a surface. The fingertip's nail was marked with nail paint to facilitate image processing and compute the position of the fingertip in image coordinates. The slow processing pipeline (4K images $\rightarrow$ USB serial communication $\rightarrow$ computer vision algorithms) precluded recording at high speed. The sampling rate was 15~Hz. To detect the fingertip's position each image was first binarized via a HSV-color threshold then a blob detection algorithm was applied to find the center of the nail polish blob in the image. This image processing was implemented with existing OpenCV algorithms.

\subsubsection{Load cells}
Eight load cells (Model TAL230A-10Kg, HT Sensor Technology Co. Ltd., Xian, China) were placed below the surface support, see Fig.\ref{fig:setup_pic}.c) and connected to signal conditioning units (Model HX711, Avia Semiconductor Ltd., Xiamen, China). The signals were routed by a microcontroller that streamed the values coded on 23~bits at a rate of 80~Hz.

\subsubsection{Microphone}
The directional microphone (Model 130A23, PCB Piezotronics Inc, Depew, NY, USA) had a sensitivity of 14~mV/Pa and was mounted on the same structure holding the cameras. The microphone pointed towards the centre of the texture to optimise sound capture.

\subsubsection{Accelerometers}
The three accelerometers (Model 8730A, Kistler Instrumente AG, Winterthur, Switzerland) had a sensitivity of 10~mV/g and were mounted along three orthogonal axes on an adjustable ring. They measured the vibrations propagating through the finger during sliding.

\subsubsection{Signal conditioning}
The accelerometers and the microphone were connected to a 4-channel laboratory amplifier and acquisition device (Model 5165A,  Kistler Instrumente AG, Winterthur, Switzerland) sampling with 32~bits resolution. The audio and acceleration signals were recorded at 12.5~kHz. The synchronisation of these signals was critical. To ensure a precise and accurate temporal alignment, the synchronisation of the microphone and the accelerometers was managed directly by the signal conditioning unit at a hardware level. The data was then sent via a TCP socket to the main computer.

\subsection{Surfaces}
We selected ten surfaces from over the one hundred samples described in \cite{bergmann_tiest_analysis_2006}, subjectively representing a diversity of materials and surface topographies.

\begin{figure}[ht!]
    \centering
    \includegraphics[width=0.5\textwidth]{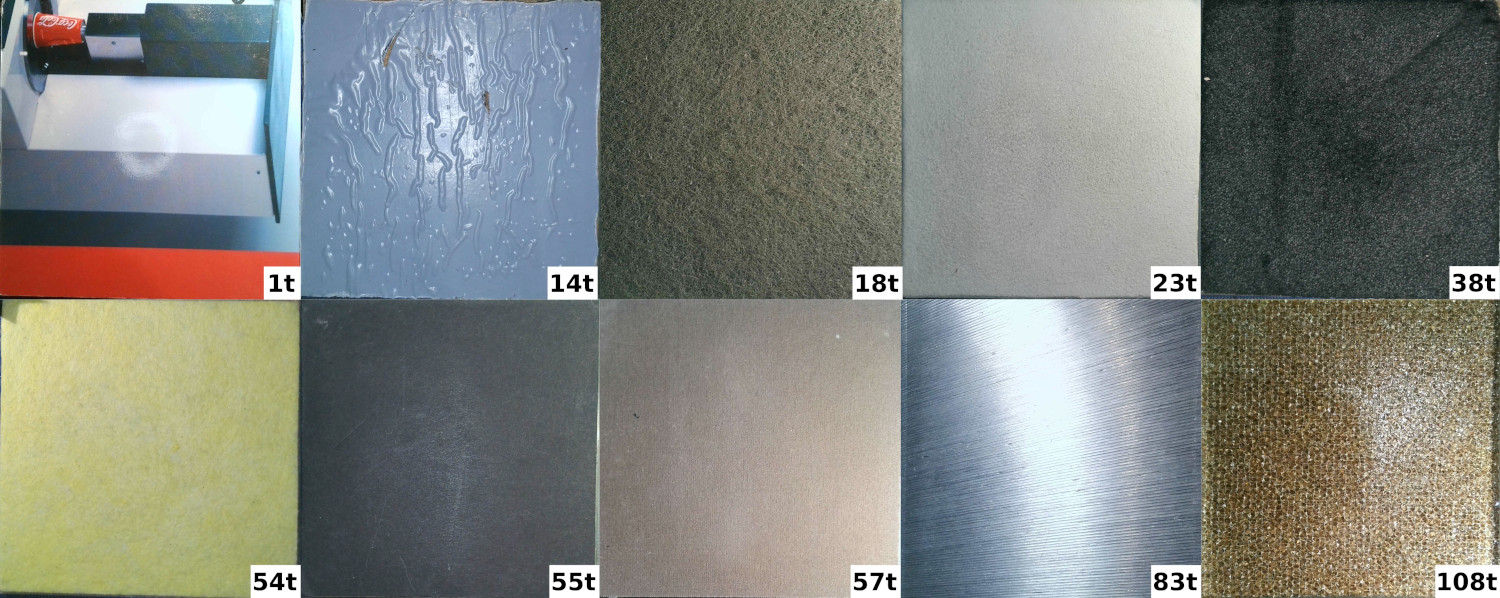}
    \caption{Images of the ten surfaces selected from~\cite{bergmann_tiest_analysis_2006}. 1t: Glossy cardboard, 14t: Plastic film textured by trapped air bubbles, 18t: Insulation fibres, 23t: Leather chamois, 38t: Plastic foam, 54t: Kitchen cloth, 55t: Thick paper, 57t: Sanding cloth Sianor J P 180, 83t: Ribbed aluminium block and 108t: Glass with medium structure.}
    \label{fig:textures}
\end{figure}

\subsection{Recording protocol}
In each trial, the volunteer sat in front of the acquisition setup and waited for the start of the recording. At the beginning of each trial, the finger was placed above the texture without touching it, enabling the calibration of the load measurement. 

To provide variability in the recorded movements, each trial included,
\begin{itemize}
    \item 10~s of lateral back and forth sliding. 
    \item 10~s of proximal/distal back and forth sliding. 
    \item 10~s of clockwise circular sliding.
    \item 10~s of anti-clockwise circular sliding.
    \item 20~s of unconstrained sliding.
\end{itemize}

The volunteer made best efforts to keep the same speed and applied load during the constrained motion phase. During the last 20~s of unconstrained motion, the volunteer was free to vary the load and the scanning speed.

\subsection{Data processing}

\begin{figure}[ht!]
\centering
\includegraphics[width=0.5\textwidth]{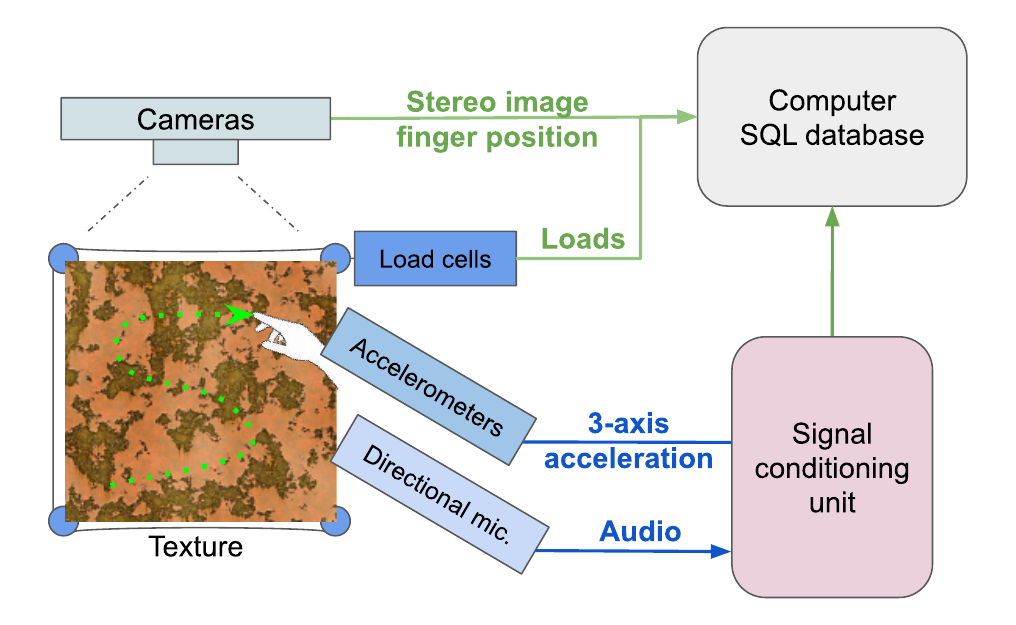}
\caption{Schematics of the recording setup. Cameras recorded images of the material and tracked the fingertip. Load cells measured forces and torques applied by the finger to the surfaces. Accelerometers measured vibrations propagating through the finger. Microphone recorded the sounds produced by the interaction of the fingertip with the surface.}
\label{fig:setup_sch}
\end{figure}

\subsubsection{Synchronisation}
To ensure the synchronisation of each source of data, the computer kernel clock was used to add a timestamp to every chunk of data received. Each chunk was then streamed via a \texttt{LabStreamingLayer} architecture~\cite{kothe_lab_2014}. In parallel, the computer ran a multi-threaded process receiving all the data and feeding it to a PSQL database. The audio and acceleration streaming at 12.5~kHz generated a lot of data. To increase the database storing speed, we used the \texttt{TimeScaleDB} PSQL extension which is specialised for time series storing.

\subsubsection{Image cropping}
Although the cameras were positioned in close proximity to the surfaces, the captured images contained part of the setup holding the studied piece of texture. Images were cropped to keep only the region of interest centred on the texture. The new origin was subtracted from position signals to maintain the position values in image coordinates.

\subsubsection{Upsampling}
The loads and positions were recorded at a lower sampling rate than the audio and acceleration signals. Force and position signals were up-sampled to 12.5~kHz to facilitate future data processing.

\subsubsection{Load representation}
The surface samples had different masses. The initial load readings were subtracted from subsequent load signals. The load vector signals, noted $f=(f_1, .., f_8)\in\mathbb{R}^8$, were mapped via least-squares optimisation to a force and torque vector $F=(f_x,f_y,f_x,\tau_x,\tau_y,\tau_z)_\mathscr{B}$ represented in Cartesian space with origin at the image centre. 


\section{Results} \label{Results}

It was previously shown in the audio domain that classification was advantageously carried out with a set of features termed ``Mel Frequency Cepstral Coefficients'' (MFCC)~\cite{vimal_mfcc_2021}. A combination of feature and machine learning classifiers, such as the Support Vector Machine (SVM) or the Random Forest algorithm, allowed the authors to classify human emotions as audio classes. A similar approach was practiced in~\cite{rong_audio_2016} where traditional machine learning classifiers were applied over a data set to perform classification.



\subsection{Classification through stereoscopic images}
\begin{figure}[ht!]
\centering
\includegraphics[width=0.5\textwidth]{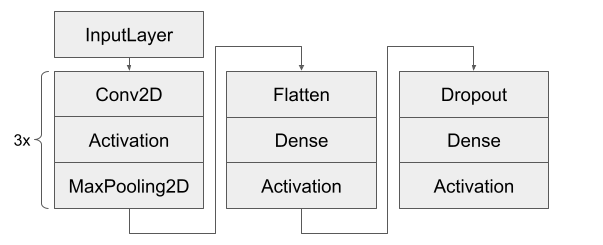}
\caption{Neural network model for image classification}
\label{fig:network}
\end{figure}

Image data augmentation was performed before image data was passed through a classification networks~\cite{zhu_texture_2018}. This step enhanced the quantity of data set and therefore, the classification performance was also improved. The classification of raw surface images was then performed using a neural network model, a simple 2D image classification model ~\cite{oshea_introduction_2015} shown in Fig.\ref{fig:network}. This network extracted the edges and contours of the image to identify similar features and correlated them within the same class. This step led to efficient differentiation in different classes. The training and test data set were respectively split into a 70:30 ratio.

\begin{figure}[ht!]
\centering
\includegraphics[width=0.4\textwidth]{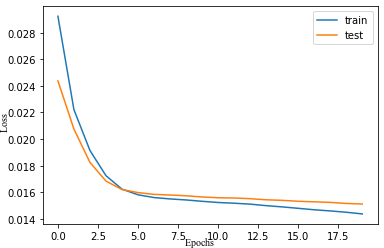}
\caption{Loss curve for image classification}
\label{fig:curve}
\end{figure}

The neural network performed well on the image data set with 87\% accuracy for training data and 83\% for test data. The loss curves in Fig.\ref{fig:curve} exhibits successful learning. The training and test curves came close to each other, which is a sign of absence of over-fitting and under-fitting. The low training loss and test loss of 0.143 and 0.157, respectively, show the good performance of this classification model.

\subsection{Classification using audio signals}

The raw audio data collected by the microphone contained redundant information. To get rid of redundant information and improve classification performance, it was essential to extract features in the raw data. Efficient features for audio classification were described in \cite{vimal_mfcc_2021}. It was shown that the features such as Mel Frequency Cepstral Coefficient, Spectral roll-off, Chromagram, Pitch, Spectral Centroid, spectral bandwidth, and RMS energy worked well for discriminating between different audio classes. 

The data set was split into training and test data sets with a ratio of 70:30. We aimed to show that the audio signals recorded by the setup were sufficient to discriminate between the selected surfaces. To investigate the performance of classification models on audio data, several standard Machine Learning models were applied and their performance was assessed through the following parameters: Classification Accuracy, Precision, Recall, and F1 Score.

Table~\ref{tab:audio_classifiers} shows the performance characteristics of each classifier applied to the audio data set. The Random Forest classifier performed the best for discriminating different classes of friction-induced sounds, leading to an accuracy of 71\%.

\begin{table}[ht!]
\begin{center}
\caption{Performance of audio data classifiers}
\label{tab:audio_classifiers}
\begin{tabular}{lllll}
Classifiers    & Accuracy(/\%) & Precision & Recall & F1 score\\
SVC            & 58.82  & 0.59  & 0.58 & 0.58         \\          
Decision Tree  & 52.94  & 0.54  &0.51  & 0.52          \\
Random Forest  & 70.59  & 0.74 & 0.7 & 0.71 \\ 
\end{tabular}
\end{center}
\end{table}

\subsection{Classification using acceleration signals}

The acceleration data captured by the three accelerometers was also split in a 70:30 ratio of training data and testing data. The feature sets which efficiently characterise friction-induced vibratory signals is still an open question. Since there are similarities between friction-induced vibratory signals and friction-induced sounds, we used the feature set used for audio Machine Learning classification models, here applied to the acceleration signals. To this end, we had to project the three-dimensional vector of acceleration onto a scalar. In the absence of a well defined optimisation criterion, we projected the acceleration vectors onto their Euclidean norm.

The classification performances of each model are shown in Table~\ref{tab:acc_classifiers}. It is clearly seen that the Random Forest classifier again performed well in assessing the differences between different surfaces.

\begin{table}[ht!]
\begin{center}
\caption{Performance of acceleration data classifiers}
\label{tab:acc_classifiers}
\begin{tabular}{lllll}
Classifiers    & Accuracy(/\%) & Precision & Recall & F1 score\\
SVC            & 61.20   & 0.61 & 0.61 & 0.60      \\
Decision Tree  & 72.23   & 0.72 & 0.72 & 0.72      \\
Random Forest  & 76.47   & 0.89 & 0.76 & 0.79       \\
\end{tabular}
\end{center}
\end{table}

\subsection{Multi-modal Classification}

\paragraph{Classification through audio and acceleration signals}
Next, we investigated classification performance when audio, and acceleration data were combined. Standard machine learning models were applied to a fusion of audio and haptic data after feature extraction. The fusion model boosted the classification accuracy of the classifiers to a great extent as shown in Table~\ref{tab:acc_audio_model}.

\begin{table}[ht!]
\begin{center}
\caption{Performance of audio and acceleration data classifiers}
\label{tab:acc_audio_model}
\begin{tabular}{lllll}
Classifiers    & Accuracy(/\%) & Precision & Recall & F1 score\\
SVC            & 80.00 & 0.84 & 0.81 & 0.80           \\
Decision Tree  & 86.60 & 0.86 & 0.86 & 0.86           \\
Random Forest  & 96.67 & 0.97 & 0.96 & 0.96   \\
\end{tabular}
\end{center}
\end{table}

\paragraph{Classification using acceleration and load signals}
In addition to accelerations, we also collected the load applied by the finger during sliding since the load applied informs about the manner users explored the surfaces. Classification was performed on load data combined with acceleration data. Table~\ref{tab:acc_force_classifier} shows the results. There was an overall enhancement in the ability of the classifiers to discriminate textures. Interestingly, the performance of multi-modal load-acceleration classification surpassed the performance of a single-modality classification.

\begin{table}[ht!]
\begin{center}
\caption{Performance of acceleration and load data classifiers}
\label{tab:acc_force_classifier}
\begin{tabular}{lllll}
Classifiers    & Accuracy(/\%) & Precision & Recall & F1 score\\
SVC            & 60   & 0.61 & 0.60 & 0.60        \\
Decision Tree  & 67.20   & 0.69 & 0.67 & 0.66        \\
Random Forest  & 82    & 0.85 & 0.82 & 0.82        \\
\end{tabular}
\end{center}
\end{table}

\paragraph{Classification using audio, acceleration, and force signals}
A fusion model of audio data with load and acceleration data did not enhance performance much. This result can be explained after comparing the performance of the acceleration/audio classifier with that of that of the acceleration/load classifier. Load data contributed less to the classification process than audio data, see Table~\ref{tab:acc_audio_force_model}.

\begin{table}[ht!]
\begin{center}
\caption{Performance of the audio, acceleration and load data classification}
\label{tab:acc_audio_force_model}
\begin{tabular}{lllll}
Classifiers    & Accuracy(/\%) & Precision & Recall & F1 score\\
SVC            & 82 & 0.84 & 0.82 & 0.82           \\
Decision Tree  & 86.6 & 0.89 & 0.86 & 0.86           \\
Random Forest  & 97.3 & 0.97 & 0.96 & 0.96   \\
\end{tabular}
\end{center}
\end{table}

\subsection{Summary}

\begin{figure}[ht!]
    \centering
    \includegraphics[width=0.5\textwidth]{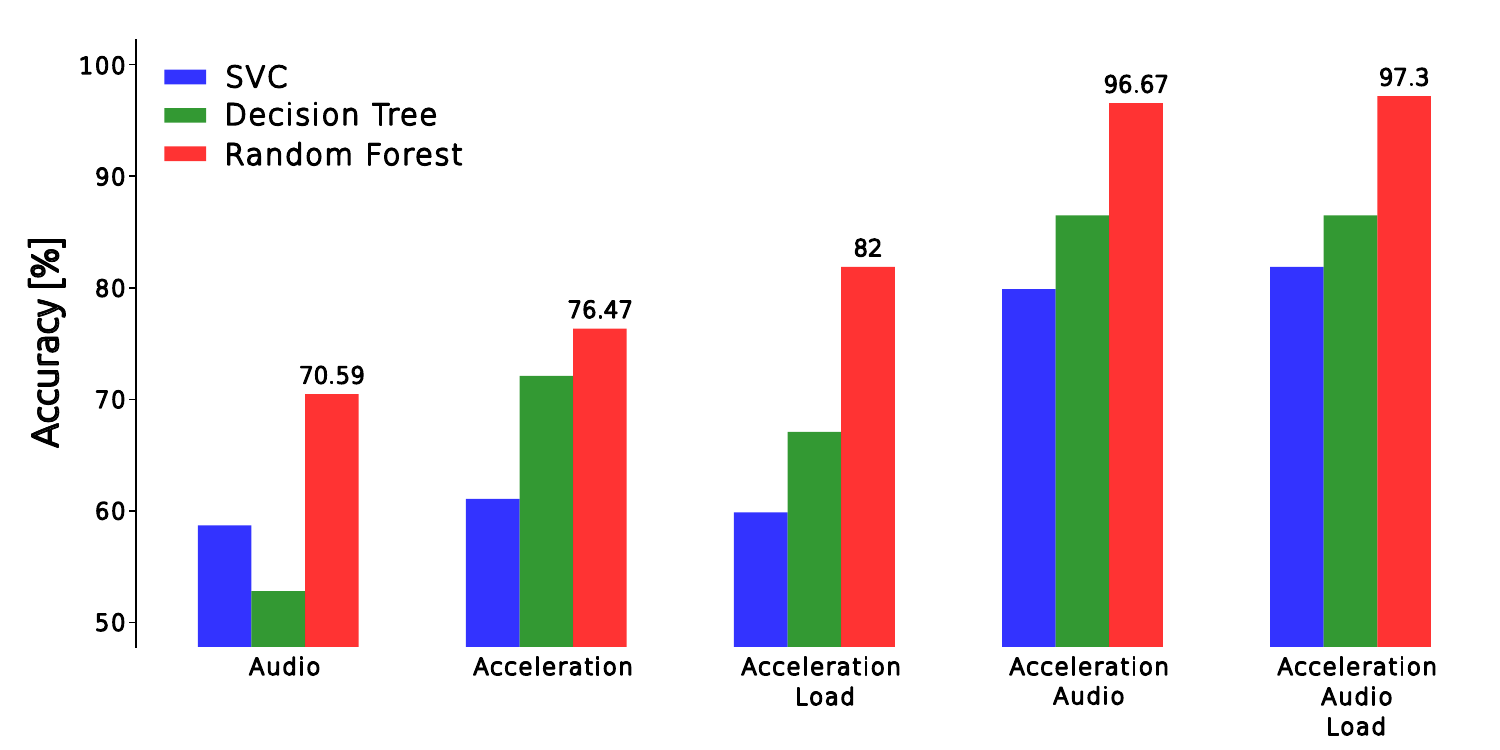}
    \caption{Accuracy of the three classifiers for each modalities and their combinations.}
    \label{fig:accuracy_comparison}
\end{figure}

Fig.~\ref{fig:accuracy_comparison} summarises the classification results. It can be observed that Random Forest algorithm persistently performed better than other classification models for audio, acceleration, acceleration/load, audio/acceleration and audio/acceleration/load classification models. Interestingly, multi-modal data classification works better than single modal data classification. There was a significant increase in accuracy of acceleration classifier when audio data was introduced. Except audio data classification, a similar pattern was observed in the performance of all classifiers (SVC, decision trees, Random Forest) where classification accuracy by SVC is lesser than decision trees which in turn was outperformed by Random Forest model. 

\section{Discussion} \label{Discussion}

We built a rich data set of multi-modal sensory signals associated with exploring with bare finger surfaces made of different materials and having varied topographies. This data set will be made publicly accessible. To assess the quality of this data set, we performed a series of tests with different types of machine learning classification algorithms. In a classification problem where the dataset is unstructured or semi-structured and not clearly understood, Support Vector Classifiers(SVC) aid such classification and are also memory efficient. The choice to use decision trees and random forest model was made because of their ease of interpretability and visualisation. However, Random forest surpasses performance as it prevents over-fitting on dataset which is a huge advantage in a classification problem.

Image, audio and haptic classification performed well in spite of the unconstrained movements of the volunteer. Classification was possible with haptic signals, audio signals, and image data. The performance of the audio classifier was below that of the haptic and image classifiers. The sounds emitted during exploration were often insufficient to discriminate between surfaces. Different surfaces gave rise to highly variable sound levels since certain surfaces produced almost no sound when scanned with a real fingertip. This result would likely be different had a rigid probe been used to scan the surfaces. 

Overall, the best performance was achieved by the Random Forest classifier compared to  other classifiers. Random Forest classifiers make use of ensemble learning methods which work through a multitude of decision trees. Each mutually exclusive branch represents a subcategory of input features. The Random Forest algorithm was superior to the Decision Tree algorithm. The Random Forest algorithm includes bootstrap aggregation which decorrelates the decision trees corresponding to different training sets. Hence, over-fitting was prevented through voting to predict an output. This strategy was very successful in the classification of different surfaces.

Further work will involve data collection with multiple volunteers in order to understand the variability in the haptic and sound signals. A comparison of signals from different participants may help us understand how humans perceive textured surfaces in spite of great variability of the available signals and the strategies they use to achieve versatile  and reliable perception.

\bibliography{references} 
\bibliographystyle{ieeetr}
\end{document}